\documentclass{ieeeconf}

\usepackage{times}
\usepackage{epsfig}
\usepackage{graphicx}
\usepackage{amsmath}
\usepackage{amssymb}
\usepackage[utf8]{inputenc}
\usepackage{overpic}
\usepackage{multirow}
\usepackage{xcolor}

\newcommand{\todo}[1]{}
\renewcommand{\todo}[1]{\textcolor{red}{TODO {#1}}}

\newcommand{\name}{LRPD}

\title{\LARGE \bf
\name : Long Range 3D Pedestrian Detection\\ Leveraging Specific Strengths of LiDAR and RGB
}

\author{Michael Fürst$^{1}$ and Oliver Wasenmüller$^{1}$ and Didier Stricker$^{1}$\\
$^{1}$DFKI - German Research Center for Artificial Intelligenc\\
        {\tt\small firstname.lastname@dfki.de}%
}

\begin{document}

\maketitle
\thispagestyle{empty}
\pagestyle{empty}


\begin{abstract}

While short range 3D pedestrian detection is sufficient for emergency breaking, long range detections are required for smooth breaking and gaining trust in autonomous vehicles.
The current state-of-the-art on the KITTI benchmark performs suboptimal in detecting the position of pedestrians at long range.
Thus, we propose an approach specifically targeting long range 3D pedestrian detection (LRPD), leveraging the density of RGB and the precision of LiDAR.
Therefore, for proposals, RGB instance segmentation and LiDAR point based proposal generation are combined, followed by a second stage using both sensor modalities symmetrically.
This leads to a significant improvement in mAP on long range compared to the current state-of-the art.
The evaluation of our \name~approach was done on the pedestrians from the KITTI benchmark.

\end{abstract}

\section{Introduction} \label{sec:introduction}

Vulnerable Road User (VRU) detection is a major challenge to be solved for autonomous vehicles.
The vehicle should avoid all obstacles including pedestrians.
To successfully achieve this, a dependable 3D detection is required.
Despite visually appealing results, current state-of-the-art approaches show suboptimal performance when numerically evaluated.
One key challenge is the detection of pedestrians at long ranges.
Between $20$ to $50$ meters is a typical breaking distance in cities at $0.3g$ breaking acceleration, leading to a comfortable driving experience.
Beyond comfort, long range detection also gives an edge on security, since it increases the amount of time available to decision making and temporal fusion.

Even though long range 3D pedestrian detection is of significant importance for autonomous vehicles and automated driving functions, there has been little research done on that topic yet.
We are the first to have an in-depth look at the performance of existing state-of-the-art approaches at long range.
Comparing their mAP at long ranges, we found strategies to improve it, by using precise RGB segmentation in combination with LiDAR distance information.
Other approaches do not have this strong focus on RGB in their proposals, even though RGB contains more information on far away objects.
Our analysis goes beyond the official KITTI benchmark~\cite{kittibenchmark} and we show that the approaches have different performance characteristics, while achieving the same KITTI score.


\begin{figure}[t]
    \begin{center}
	    \begin{overpic}[width=0.98\linewidth,,tics=10]{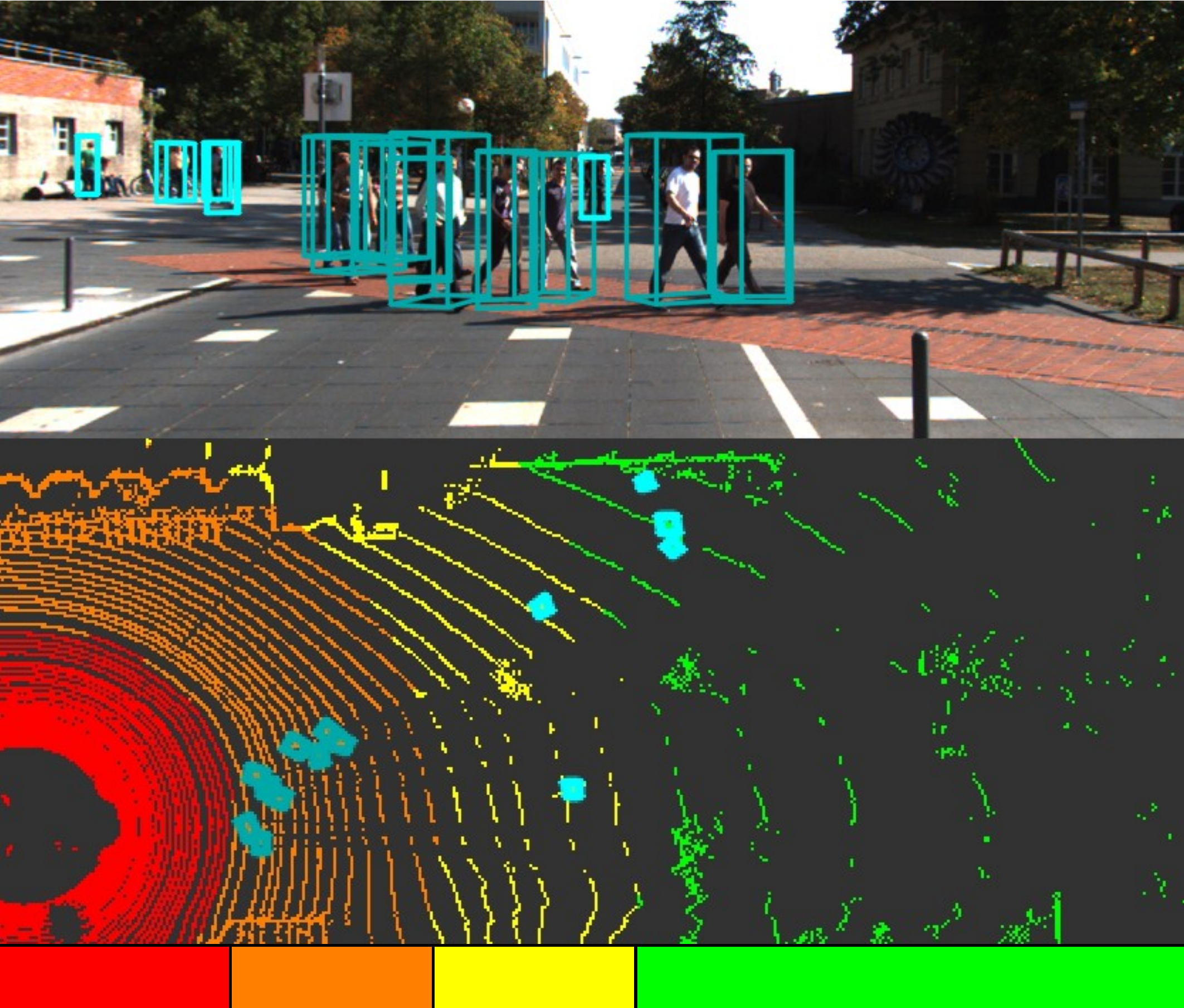}
		    \put(0.5,1){immediate}
   		    \put(24,1){short}
   		    \put(38,1){medium}
   		    \put(58,1){long (30+ m)}
		\end{overpic}
    \end{center}
    \caption{Immediate (0-10m), short (10-20m), medium (20-30m) and long range ($\geq$ 30m) can be used to analyse a detectors performance. Recognizing its importance, we focus on long range and outperform the current state-of-the-art.}
    \label{fig:few_points}
\end{figure}
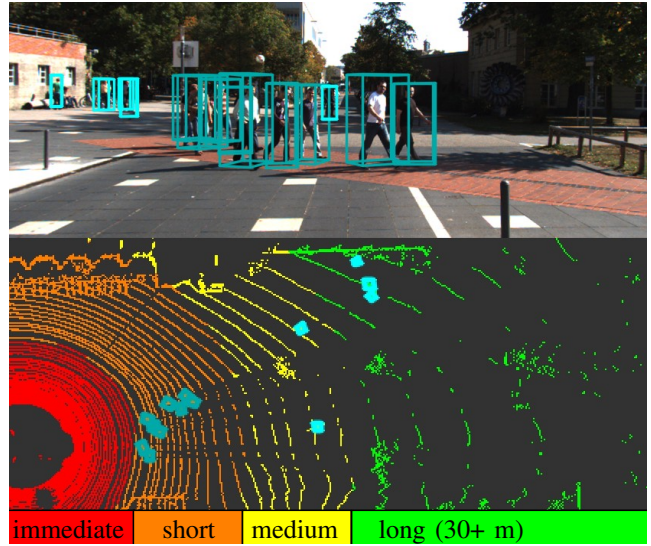

Since all existing approaches have deficits in long range 3D pedestrian detection, we propose:
\begin{itemize}
	\item A new proposal generation stage, leveraging density of RGB images and precision of LiDAR data, followed by a symmetrical refinement stage,
	\item a distance based proposal augmentation schema to further enhance the performance of algorithms,
	\item an evaluation of existing approaches and our \name~approach with a significant improvement in mAP over the current state-of-the-art on long range pedestrian detection.
\end{itemize}


\section{Related Work} \label{sec:related_work}

3D detection of objects is a key technology required for autonomous driving and automated driving functions.
Recently, there have been significant advances in this field.

PointPillars~\cite{pointpillars} uses only LiDAR information in an SSD~\cite{ssd} like network architecture.
One key contribution of PointPillars is the introduction of pillars instead of voxels as in VoxelNet~\cite{voxelnet}.
A feature embedding is computed for every non empty pillar, which is then used in a 2D BEV feature map, enabling efficient 2D convolutions.
The 2D convolution backbone is an FPN~\cite{fpn} which is then used for an SSD style decoder head.
We consider PointPillars as LiDAR only single-stage 3D detector.

Most approaches however are multi stage approaches, using an ROI-pooling approach derived from the R-CNN series~\cite{fastrcnn, rcnn, fasterrcnn}.
F-PointNet~\cite{fpointnet} leverages the great performance of 2D detectors, by predicting 2D bounding boxes, which are used to define a frustum in 3D space.
All points from the LiDAR, which fall inside the frustum, are then used in a PointNet~\cite{pointnet} which predicts an instance segmentation of the points.
Only points which are segmented as foreground are then used for the final prediction of the bounding box by a PointNet.
F-PointNet relies on 2D bounding box prediction in the RGB image to reduce the search space for objects and then does the final prediction of objects based only on the frustum LiDAR pointcloud using PointNets.

Whereas F-PointNet~\cite{fpointnet} uses classical 2D proposal based ROI pooling, AVOD~\cite{avod} uses 3D ROI pooling.
The architecture is similar to a Faster-R-CNN~\cite{fasterrcnn}, but uses a static 3D anchor grid to crop and resize the features before fusing them and applying a fully connected network as an RPN on them.
The outputs of this RPN are then used to crop and resize the features again, fuse them and finally estimate the bounding box by using a fully connected layer.
AVOD is a two-stage 3D detector using RGB and LiDAR symmetrically in the RPN and refinement stage.

In contrast to previous approaches, IPOD~\cite{ipod} uses an unconventional proposal strategy focused on semantic segmentation.
RGB segmentation is used to find the relevant points in the pointcloud and samples boxes around them.
The proposal stage has influenced our proposal stage, since it can yield proposals from only one inlier point.
However, IPOD only uses LiDAR pointclouds for its refinement stage.

State-of-the-Art is typically compared on the KITTI Benchmark~\cite{kittibenchmark}.
It consists out of 7481 training images and 7518 testing images.
Containing 4487 pedestrian instances in 1779 images for the training set.
The top performing approaches are shown in Table~\ref{tab:kitti_bev}.
For our in-depth analysis we use AVOD~\cite{avod}, PointPillars~\cite{pointpillars} and F-PointNet~\cite{fpointnet}, since they are the best performing open source algorithms and represent the common algorithm categories.
Those approaches can be differentiated into single or two stage approaches, LiDAR only or RGB + LiDAR and the way how they leverage 2D information and 2D detection performance.

\begin{figure}[t]
    \begin{center}
       \includegraphics[width=0.98\linewidth]{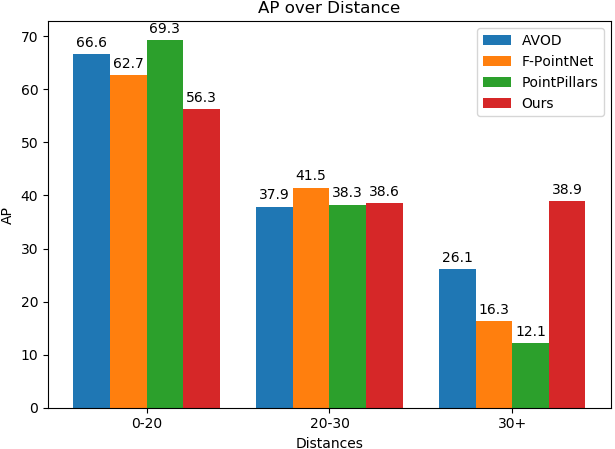}
    \end{center}
    \caption{Mean Average Precision (mAP) on KITTI BEV \textit{easy} for pedestrians in 3D plotted for different ranges, using available state-of-the-art and our \name~(with grounding) approach.
We outperform all approaches by a margin on long range (30+ m), while maintaining competitive on short and medium range.}
    \label{fig:compare_distance_easy}
\end{figure}

We analyze the performance by measuring the mAP for different ranges.
The mAP is computed by limiting the range, in which detections are considered, to an interval.
As can be seen in Figure~\ref{fig:compare_distance_easy}, the performance of approaches which have a focus on LiDAR for the final predictions (F-PointNet and PointPillars) have a significant drop in performance for long range when compared to approaches like AVOD which use both sensors.
VMVS~\cite{VMVS}, STD~\cite{STD}, IPOD~\cite{ipod}, F-ConvNet~\cite{fconvnet} and TANet~\cite{tanet} were not evaluated in detail, due to the lack of source code available or similarity to other approaches.
VMVS, F-ConvNet and TANet are conceptually similar to AVOD, F-PointNet or PointPillars.

By analysing the KITTI dataset~\cite{kittidataset}, we found a few indications which support our observations.
On one hand, the number of LiDAR points for objects at long range is very low.
On average an object further away than 30 meters has only 18 Points.
Challenging objects consist of 5.5 points only, which makes distinction between noise, poles and pedestrians almost impossible.
On the other hand, in the RGB image the same objects consist of on average 706 pixels, which is sufficient for a classification in most cases.
The differentiation between the pedestrian and any other pole like object can only be made by using RGB as a secondary input.
PointPillars using only LiDAR is at a clear disadvantage.
Similarly, F-PointNet relying on pointclouds for the refinement stage.
AVOD achieves the best results, what can be attributed to the usage of RGB and LiDAR throughout all stages of the network.
In other domains like scene flow estimation the potential of LiDAR and RGB fusion was already shown. \cite{lidarflow} used LiDAR to improve the matching, interpolation and consistency check by constraining each step of their pipeline with LiDAR measurements.
To the best of our knowledge there exists no algorithm, which specifically targets 3D pedestrian detection at long range.
Thus, we present for the first time an approach targeting long range pedestrian detection from sparse LiDAR points and RGB data.

\section{Methodology} \label{sec:architecture}

\begin{figure*}[t]
    \begin{center}
       \includegraphics[width=0.98\linewidth]{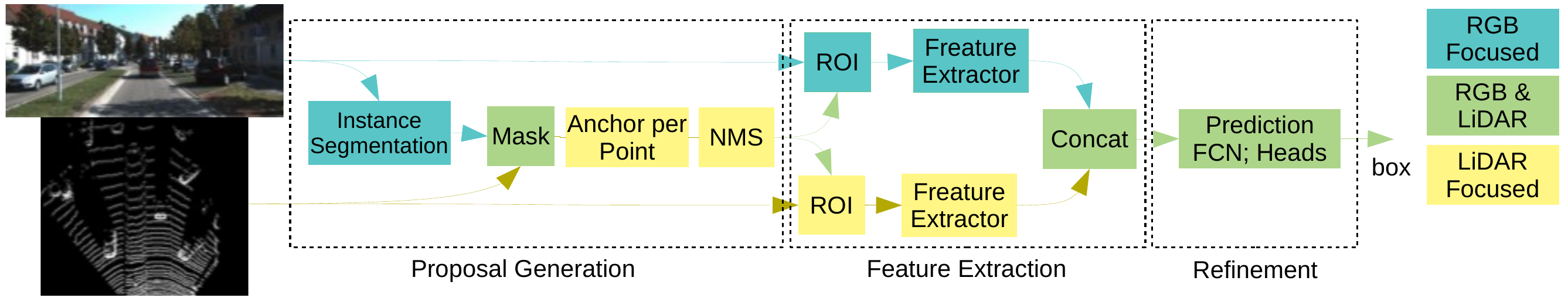}
    \end{center}
    \caption{The RGB image and LiDAR pointcloud are used for a proposal generation, then ROI cropped. The crops are used in a feature encoder, concatenated and finally the refinements for the proposal are predicted. The proposal is generated by applying Mask-RCNN instance masks to the LiDAR, creating a box around every foreground point and then using non-maximum suppression (NMS) to reduce the number of proposals.}
    \label{fig:architecture}
\end{figure*}

From our analysis we identify the low number of LiDAR points per object as a problem for long range detection.
Our approach, called \name, solves this problem.
We have discovered, that a strong focus on RGB in the proposal stage and symmetrical fusion of LiDAR and RGB only in the second stage achieves best performance on long range detection tasks.
All approaches presented in Section~\ref{sec:related_work}, are either symmetrical in both stages or do not use both LiDAR and RGB in their second stage.

Based on an instance segmentation, our approach generates a proposal.
The proposal is used for ROI cropping of the image and the pointcloud data, before feeding them through an encoder.
Finally, the two resulting feature vectors are concatenated and a fully connected layer computes the objectness and refinements (Figure~\ref{fig:architecture}).

\subsection{Proposal Generation}

Our proposal generation is a four step process inspired by IPOD~\cite{ipod}, starting with segmentation, then masking the LiDAR with those instances, assigning a box to each positive point and finally applying non-maximum suppression (NMS).
Even though our proposal pipeline looks similar to IPOD it varies in many details.

In contrast to IPOD -- which uses semantic segmentation -- we use an instance segmentation network, e.g. Mask-RCNN~\cite{maskrcnn}, which has shown strong results on KITTI and many other datasets.
Instance segmentation should be preferred over semantic segmentation for two reasons.
The latter often miss-classifies the area in-between two pedestrians, separated by a few pixels, as pedestrian.
This leads to an unnecessary large amount of positive points resulting in a lot of unnecessary proposals.
The other advantage of instance segmentation is that non-maximum suppression can be limited per instance, resulting in at least two proposals for two instances.
With semantic segmentation this cannot be ensured.

After computing the instance masks, we apply them to the pointcloud to obtain the points related to an instance.
This is done by projecting the pointcloud into the image space and assigning the points to the instance mask in which they fall or discard them if they are not inside any instance mask.
The instance information is then applied back onto the original 3D pointcloud.

Now all 3D points that have an instance mask assigned are converted into a 3D bounding box proposal.
The proposal has a width, height and length, defined by the average box on the dataset, and the x, y, z centeroid, defined by the point, used to generate it and the instance mask id is also stored.

Finally, all proposals undergo non-maximum suppression with a birds-eye-view (BEV) intersection over union (IoU) using a threshold of $0.5$.
However, our non-maximum suppression (NMS) harvests the knowledge about instances intelligently.
The score for the NMS is defined by the number of inlier points with the same instance mask id a box has.
This reduces the number of proposals greatly to 1-5 proposals per instance detected.
With these changes in the proposal generation, the step of ambiguity reduction is not necessary, since we get a small amount of proposals per instance.

If the instance segmentation detects an object and the object contains at least one point, a proposal at the correct position is generated.
Therefore, the main limiting factor of the proposal generation stage is the performance of the instance segmentation network.

\subsection{Proposal Augmentation for Training}

Besides the ability to generate proposals reliably from as little as one LiDAR point, proposal augmentation is a factor for the performance of our approach.
Since the proposal stage often generates only one correct proposal per instance, there is little variance in proposal offsets.
For the best performance of our refinement network, we increase variance by exploring various approaches for proposal augmentation.
Random 3D displacement of the boxes and \textit{grounding} has the most impact on performance.
Random displacement directly increases variance in the proposals, whereas grounding simplifies the refinement task itself, reducing the necessity for large variance in the training data.

As simple data augmentation we use random displacement.
All proposals are randomly displaced multiple times and a single copy of the original proposal is kept.
After displacement a check is done, if the proposal is still close enough to the ground truth.
This introduces what we call \textit{close negative proposals}.
Close negative proposals are close to a prediction, but not within the threshold.

At long range the proposals have less points.
Therefore, using point based sampling, they might be significantly too high or low.
To alleviate this issue we introduced what we call \textit{grounding}.
Grounding generates a pillar around the proposal.
In this context pillar we find the lowest point and assume it to be the ground.
The ground is preferred over the top, since poles stick out beyond the head of a pedestrian.

However, as objects get closer to the vehicle or are heavily occluded, the ground might not be visible and performance degrades.
This is a problem for objects in immediate range.
A simple yet effective solution is to use grounding only for non-immediate range.
Merging random displacement and grounding based on the distance is called \textit{combined} by us, since it combines two ways of augmentation.

The preference of the augmentation method depends on the use case.
For best overall performance combined is the method of choice.
For short range prediction just random augmentation is preferred and for long range prediction random augmentation with grounding yields best results.

\subsection{Box Prediction}

The proposals of Section III.B are used for the final box prediction.
They are used in a ROI cropping step on the image and the LiDAR pointcloud.
For image ROI cropping, the proposal is projected into the 2D image space.
The minimal 2D box containing all corners is then computed.
That 2D box is scaled by a factor of 1.5, increasing the context area.
Finally, the image is cropped.

The pointcloud gets cropped by using a 3D axis aligned box around the centeroid of the proposal.
Similar to RGB, increasing the size of the region crop increases the context available to the neural network.
The crop of the pointcloud is then voxelized, using a regular grid of shape 64x64x9.
Then, the points in each voxel get encoded as the highest point in a voxel and the density of a voxel.

Both encoded crops are used by two separate 2D convolutional encoders.
One operating in BEV and one in image view.
As an encoder for images DenseNet~\cite{densenet} works best. Any encoder works, however with different performances, as we show in our ablation study.
For the BEV feature representation we use a network architecture inspired by VGG16~\cite{vgg} explained in our implementation details.

The two encodings are global average pooled separately and then concatenated.
The final refinement is then done by a fully connected layer with a ReLU activation and four parallel fully connected layers with different activation functions.
The first output regresses a single value with a sigmoid activation function representing the objectness.
The second output regresses three values for the position offset $\Delta x, \Delta y, \Delta z$ from the proposal box.
The size offsets $\Delta l, \Delta w,\Delta h$ are regressed as another set of three values.
The last regression head predicts the orientation as two outputs with tanh activation function.
The orientation needs two values for $s_\theta = \sin(\theta)$ and $c_\theta = \cos(\theta)$.

\subsection{Loss}

The first component of our loss is for the objectness of an anchor.
As the number of positive and negative proposals is unbalanced, focal loss~\cite{retinanet} is a good choice.
However, we do not know the optimal $\gamma$ for our use case, therefore, we use the automated focal loss \cite{automated_focal_loss} for objectness, defined as
\begin{equation}
    L_{objectness} = - ( 1 - p_t )^{-\log(\hat{p}_{t})} \log(p_t),
\end{equation}
where $p_t$ is the probability of the correct prediction and $\hat{p}_{t}$ the expected probability of a correct prediction as defined in~\cite{automated_focal_loss}.
Then, for the regression of the centeroid ($xyz$) and the size ($lwh$) common smooth l1 loss is used as $L_{xyz}$ and $L_{lwh}$.
The heading $\theta$ which is encoded as $s_\theta, c_\theta$ has a smooth l1 loss on the encoding defined as
\begin{equation}
    L_{\theta}
    = \texttt{smooth\_l1}\left(
    \begin{pmatrix}
    s_\theta\\
    c_\theta
    \end{pmatrix},
    \begin{pmatrix}
    \sin(\theta_{gt})\\
    \cos(\theta_{gt})
    \end{pmatrix}
    \right),
\end{equation}
where $\theta$ is the rotation around the up axis.
Finally, all losses are summed, yielding best results without weighing.
The total loss for backpropagation is defined as
\begin{equation}
    L = L_{xyz} + L_{lwh} + L_{\theta} + L_{objectness}
\end{equation}
where $L_{xyz}$, $L_{lwh}$, $L_{\theta}$ and $L_{objectness}$ are as defined above.

\subsection{Implementation Details}

Our BEV encoder consists of 5 conv-pool-blocks with two 3x3 convolution and batch norm followed by a 2x2 max pooling.
The number of filters for the layers are [24, 48], 2x [32, 64], 2x [16, 32].
After that there are two 1x1 convolution layers reducing the feature size to 16 and finally to 10.
We chose such low parameter numbers to counter overfitting on the KITTI pedestrian dataset.

\name~is trained using adam optimizer with $\beta_1 = 0.9$, $\beta_2 = 0.999$ and $\epsilon = 10^{-8}$.
Our learning rate follows an exponential decay schedule, starting at $10^{-4}$ and ending at $10^{-6}$ after 10 epochs of training using a batch size of 32.
For regularization we use batch normalization and orthogonal weight initialization inside our network combined with l2 weight decay at a rate of $10^{-3}$.
The encoders are used in the ImageNet~\cite{imagenet} pretrained version provided by \texttt{tf.keras}.

For further regularization we use data augmentation.
Principal component jitter, left to right flipping, micro translations (0-3px offset) and resolution change (via scaling) are applied on the images.
For LiDAR jitter on the xyz points (white noise) is added and a micro translation (0-3cm) is applied to the entire pointcloud.
This data augmentation is just distortion of the input and does not replace the proposal augmentation in Section~III.B.
We apply data augmentation nine times including the augmentation from Section~III.B,
increasing our training dataset ten fold adding significant variance.

Overall, we have developed \name, a specialized approach for long range pedestrian detection, starting with a proposal stage and using RGB instance masks for 3D LiDAR point based proposals.
Then, the proposals are refined, predicting the position, size, orientation and objectness in a second stage.
To further improve performance proposal augmentation is introduced and automated focal loss is applied.
\section{Evaluation} \label{sec:evaluation}

We evaluated our \name~network architecture on the KITTI benchmark focusing on birds-eye-view (BEV), since it is most important for autonomous vehicles with road users moving on the ground plane.
The KITTI BEV metric uses top-down IoU to classify correct predictions with a threshold of $0.5$ for pedestrian IoU. Furthermore, we inspected performance using precision, recall, f1-score, mean error and mAP with distance from center as a criterion.

\subsection{Results}

\begin{table*}
\begin{center}
\begin{tabular}{|l|c|c|c|c|c|c|}
\hline
\multirow{2.3}{*}{\textbf{Method}}
    & \multicolumn{2}{c|}{\textbf{Moderate}}
        & \multicolumn{2}{c|}{\textbf{Easy}}
            & \multicolumn{2}{c|}{\textbf{Hard}}     \\
    \cline{2-7}
    & \textbf{all}
        & \textbf{long range}
        & \textbf{all}
        & \textbf{long range}
        & \textbf{all}
        & \textbf{long range}                              \\
\hline\hline
AVOD~\cite{avod} & 50.32 & 12.00 & 58.49 & 26.14 & 46.98 & 12.09 \\
- VMVS~\cite{VMVS} & \textbf{50.34} & - & 60.34 & - & 46.45 & - \\
PointPillars~\cite{pointpillars} & 48.64 & 4.24 & 57.60 & 12.14 & 45.78 & 4.26 \\
- TANet~\cite{tanet} & 51.38 & - & 60.85 & - & 47.54 & - \\
F-PointNet~\cite{fpointnet} & 49.57 & 12.55 & 57.13 & 16.30 & 45.48 & 10.94 \\
IPOD~\cite{ipod} & 49.79 & - & \textbf{60.88} & - & 45.43 & - \\
MMLab-PointRCNN~\cite{pointrcnn} & 46.13 & 2.30 & 54.77 & 9.51 & 42.84 & 2.62 \\
\hline
\name~w/ Rand. Aug. (ours) & 47.04 & 15.57 & 51.94 & 33.92 & 46.51 & 15.57 \\
\name~w/ Grounded (ours) & 39.89 & \textbf{15.63} & 45.59 & \textbf{38.92} & 39.88 & \textbf{15.65} \\
\name~w/ Combined (ours) & 50.01 & 15.45 & 56.41 & 27.00 & \textbf{48.46} & 13.64 \\
\hline
\end{tabular}
\caption{The best performing approaches on KITTI pedestrian BEV mAP compared to \name~(ours).
On \textit{all} (ranges) \name~is comparable to the state-of-the-art, whereas on \textit{long range} ($>30$m) \name~outperforms all reproducible state-of-the-art approaches, not all were reproducible.
Since hard examples have slight bias towards distant objects, our \name~outperforms other state of the art approaches in this category. \name s performance can be attributed to it being designed towards long range detection.}
\label{tab:kitti_bev}
\label{tab:kitti_long_range}
\end{center}
\end{table*}

Table~\ref{tab:kitti_bev} compares \name~against the top performing open source approaches, reproducing their results from the paper.
Our approach is competitive, but KITTI BEV pedestrians benchmark weights samples by occurrence and equally not by distance range.
Whereas close objects are over represented in KITTI, the far objects have nearly no importance for the overall score on the KITTI benchmark.
However, especially long range detections are important for security and smooth maneuvers.
Therefore, we analysed how the performance changes with respect to long range.

\begin{figure}[t]
    \begin{center}
       \includegraphics[width=0.98\linewidth]{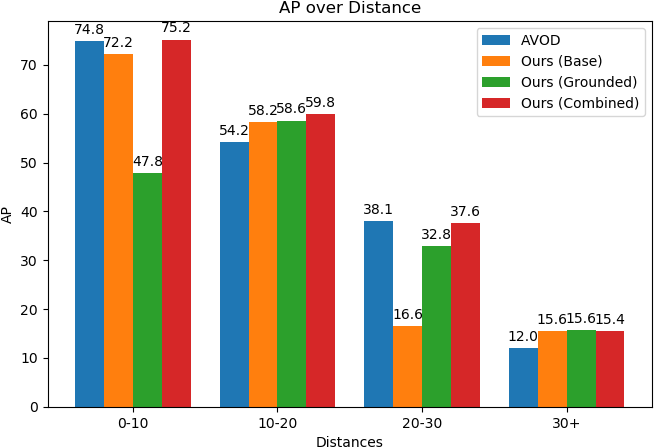}
    \end{center}
    \caption{BEV Detection mAP on KITTI \textit{moderate} for pedestrians plotted for different ranges, using AVOD as the best baseline and our approaches. While AVOD is comparable to the best configuration for immediate (0-10m) to medium range (20-30), \name~outperforms on long range (30+ m) with all configurations.}
    \label{fig:compare_distance_ours}
\end{figure}

We introduce an evaluation metric, that measures performance of object detectors at different ranges,
since detection at long range is not separately evaluated in KITTI.
To achieve this, we limit the KITTI metric to different intervals.
We chose 0-10 meters for the immediate proximity, 10-20 meters for close range, 20-30 meters for medium range and above 30 meters for long range (Figure~\ref{fig:compare_distance_ours}).
This means however, that our comparison is limited to publications with source code available.
The top performing algorithms on KITTI BEV pedestrian detection are AVOD~\cite{avod}, F-PointNet~\cite{fpointnet}, PointPillars~\cite{pointpillars} and Point-RCNN~\cite{pointrcnn}.

We observed deficits in current state-of-the-art approaches in a good mAP for short range but suffering from worse detection rates at long range.
Our \name~however is able to maintain performance for long range (30+) Figure~\ref{fig:compare_distance_ours}.
Even though none of \name s configurations can maintain performance perfectly, they outperform the other approaches as can be seen in Figure~\ref{fig:compare_distance_easy}.
For an in depth comparison at long range, Table~\ref{tab:kitti_long_range} reports our results and the results of AVOD, F-PointNet, PointRCNN and PointPillars for long range.
PointPillars which is only LiDAR based with 4.24-12.14 barely detects any object, whereas other approaches leveraging LiDAR and RGB achieve better results (10.94-26.14).
Our approach with grounding on the proposal boxes significantly outperforms all others with mAP values of 15.63-38.91.
Long range mAP was improved by a factor of 1.5 over the state of the art.

\begin{figure*}[t]
    \begin{center}
      \begin{overpic}[width=0.98\linewidth,,tics=10]{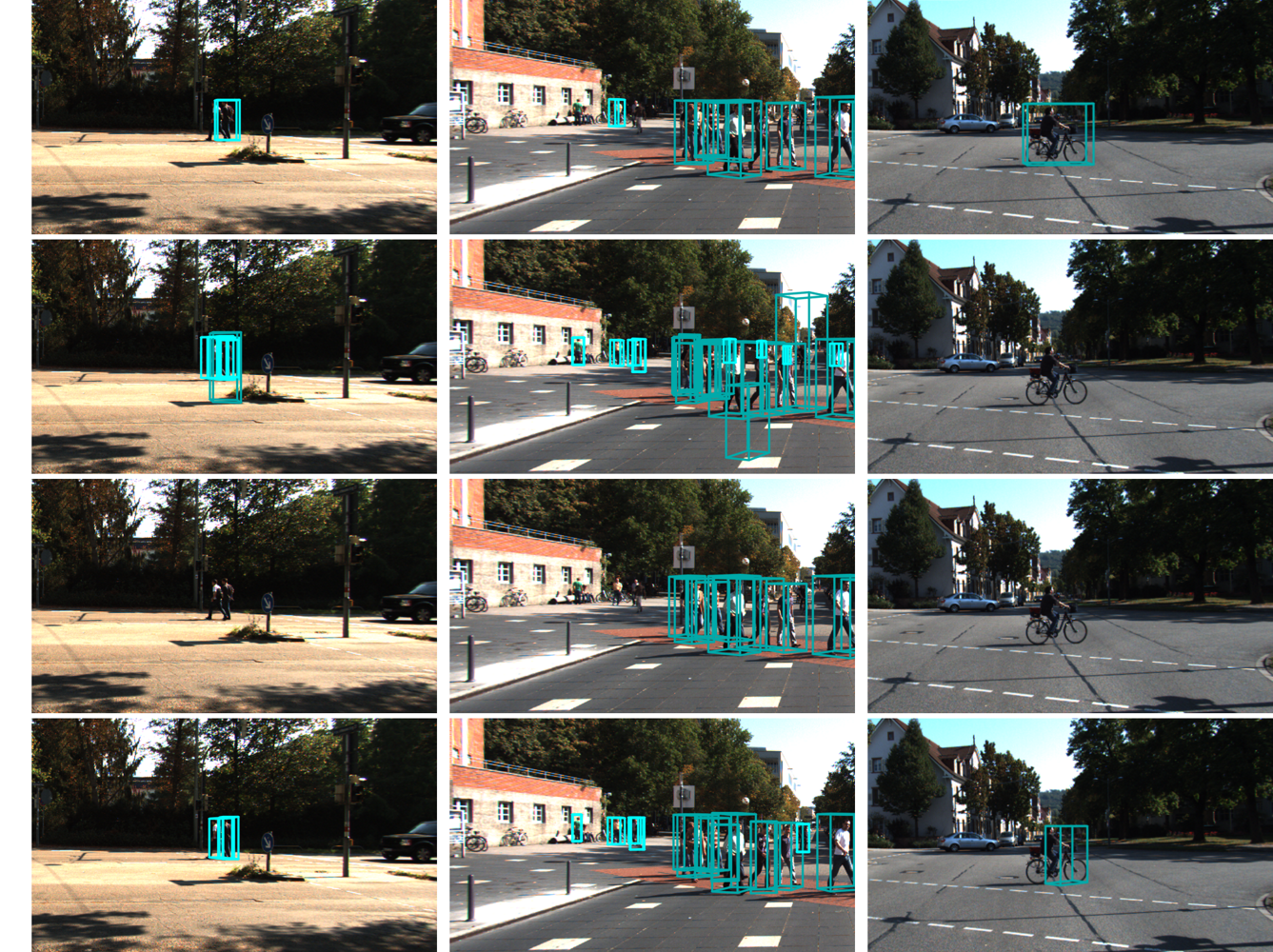}
        \put(00,61){\rotatebox{90}{AVOD~\cite{avod}}}
        \put(00,41){\rotatebox{90}{F-PointNet~\cite{fpointnet}}}
        \put(00,22){\rotatebox{90}{PointPillars~\cite{pointpillars}}}
        \put(00,04){\rotatebox{90}{\name~(ours)}}
     \end{overpic}
    \end{center}
    \caption{Qualitative comparison of the state-of-the-art to \name~(ours). 3D pedestrian detections at less than 20 meters are darker to emphasize on long range detection. The Left column shows two pedestrians. Our Approach and F-PointNet are capable of detecting both, F-PointNet however has false positives (FP). The center column shows a typical prediction, where AVOD and PointPillars are not able to detect far away pedestrians (left) and F-PointNet has a high FP-Rate when still detecting far away objects. Only \name~is able to detect the far pedestrians while maintaining precision at medium range. Finally, the right column shows a common failure case of pedestrian detectors, AVOD and ours misclassify a person on a bicycle as a pedestrian.}
    \label{fig:compare_visually}
\end{figure*}
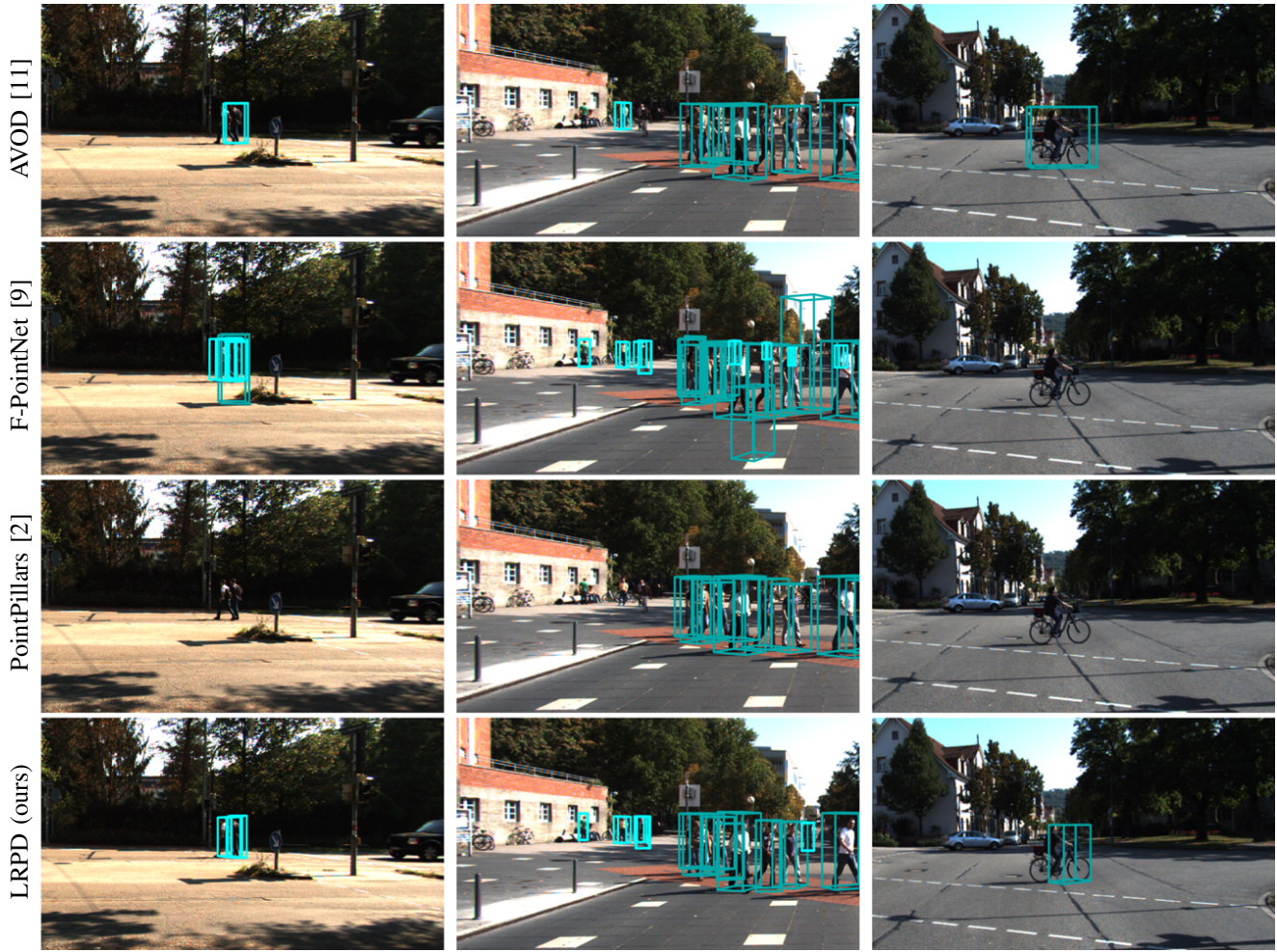

Even though all approaches have similar KITTI BEV mAP values, their failure cases are different.
In Figure~\ref{fig:compare_visually} we show some qualitative results, visualizing the typical failure modes.
When two pedestrians are close to each other and overlapping, our approach is able to reliably detect both due to the instance segmentation.
LiDAR only approaches have difficulties detecting easily visible objects at long range,
leading to PointPillars not predicting long range objects reliably.
At short to medium ranges (Figure~\ref{fig:compare_visually} center column) all approaches perform well with the difference showing on long range detections in the background.
However, since our \name~approach uses person instance masks it confuses cyclists as a pedestrian, like AVOD.
F-PointNet and Point-RCNN have very poor precision (0.22 and 0.28) at a maximum recall of 0.62 and 0.55 resulting in the numerous false positives of F-PointNet in Figure~\ref{fig:compare_visually}.
In contrast, our approach leads with 0.76 precision at a recall of 0.69 as a maximum.
However, regarding 3D euclidean error our network lacks behind other state of the art networks explaining the worse mAP values despite better precision and recall (Table~\ref{tab:kitti_detailed_eval}).
Our detailed quantitative evaluation in Table~\ref{tab:kitti_long_range} (long range), Table~\ref{tab:kitti_detailed_eval}, Figure~\ref{fig:compare_distance_easy} and Figure~\ref{fig:compare_distance_ours} is able to capture range based error effects in contrast to the original KITTI metric.

\begin{table}
\begin{center}

\begin{tabular}{|l|c|c|c|c|}
\hline
Method & F1 & Prec. & Rec. & 3D Euc. Err. \\
\hline\hline
AVOD~\cite{avod} & 0.48 & 0.57 & 0.43 & 0.14m \\
PointPillars~\cite{pointpillars} & 0.52 & 0.66 & 0.43 & \textbf{0.11m} \\
F-PointNet~\cite{fpointnet} & 0.33 & 0.22 & 0.62 & 0.16m \\
MMLab-PointRCNN~\cite{pointrcnn} & 0.37 & 0.28 & 0.55 & 0.12m \\
\hline
\name~w/ Rand. Aug. (ours) & 0.70 & 0.70 & 0.69 & 0.17m \\
\name~w/ Grounding (ours) & 0.71 & 0.68 & 0.74 & 0.22m \\
\name~w/ Combined (ours) & \textbf{0.73} & \textbf{0.76} & \textbf{0.69} & 0.18m \\
\hline
\end{tabular}
\caption{Precision and recall (at best F1-score) using 1 m euclidean distance as error threshold for assignment}
\label{tab:kitti_detailed_eval}

\end{center}
\end{table}

\begin{table}
\begin{center}
\begin{tabular}{|l|c|c|c|c|}
\hline
Proposals & Encoder & Refinement Modality & BEV mAP \\
\hline\hline
Baseline & VGG16~\cite{vgg} & LiDAR + RGB & 30.2 \\
Baseline & VGG19~\cite{vgg} & LiDAR + RGB & 39.9 \\
Baseline & ResNet50~\cite{resnet50} & LiDAR + RGB & 33.4 \\
Baseline & Xception~\cite{xception} & LiDAR + RGB & 31.3 \\
Baseline & MobileNet~\cite{mobilenet} & LiDAR + RGB & 28.5 \\
\hline
Baseline & DenseNet~\cite{densenet} & LiDAR + RGB & 43.4 \\
Baseline & DenseNet~\cite{densenet} & LiDAR & 30.9 \\
Baseline & DenseNet~\cite{densenet} & RGB & 0.3 \\
\hline
Rand. Aug. & DenseNet~\cite{densenet} & LiDAR + RGB & 51.1 \\
Grounding & DenseNet~\cite{densenet} & LiDAR + RGB & 43.5 \\
\textit{Combined} & \textit{DenseNet~\cite{densenet}} & \textit{LiDAR + RGB} & \textbf{54.8} \\
\hline
\end{tabular}
\caption{In our ablation studies we show the impact of different design decisions on the performance of our network. The final \name~configuration is marked in italic.}
\label{tab:kitti_ablation}

\end{center}
\end{table}

\subsection{Ablation Studies}

To verify the effectiveness of all components in our approach, we derived ablation studies.
These were performed on the KITTI pedestrian BEV benchmark regarding moderate AP.
We swapped out the encoder, changed the input modality and the proposal augmentation separately to observe the impacts.
The ablation studies were done on a 80:20 split on the KITTI Benchmark different from the 50:50 split used for our comparison with other approaches to avoid a bias.
In our ablation studies (Table~\ref{tab:kitti_ablation}) we show that both LiDAR and RGB are integral parts to the success of our approach.
By removing RGB or LiDAR from the second stage, described in Section~III.C, the performance degrades significantly, highlighting the importance of using both sensor modalities.
Furthermore, the encoder choice has as strong impact on performance of the network. Where VGG19~\cite{vgg} and DenseNet~\cite{densenet} significantly increase the performance, MobileNet~\cite{mobilenet} slightly decreases the performance.

When varying the proposal method described in Section~III.B, we can show that the choice of proposal augmentation has a significant impact on mAP (up to +11.4).
Baseline describes creating the proposals without any data augmentation and achieves the worst results of our proposal configurations.
Even though random augmentation has a far better overall moderate BEV mAP (Table~\ref{tab:kitti_ablation}) grounding yields better results for medium and long range detection (Figure~\ref{fig:compare_distance_ours}).
Combining the strong performance of random augmentation for short range detection with grounding for medium and long range detection, as described in Section~III.B, yields best overall results, with only 0.1 mAP reduction on long range.
Despite a low overall BEV mAP of 43.5 grounding achieves the best results on long range (Table~\ref{tab:kitti_bev}).
This proves that best mAP on KITTI does not correlate to detecting pedestrians better at long range.
Therefore, our distance range based analysis of the mAP is mandatory to gain insights into the performance characteristics of approaches,
unveiling that on long range detection \name~significantly outperforms other state-of-the-art approaches.

\section{Conclusions} \label{sec:conclusions}

In our analysis, we proved limitations of the state-of-the-art in long range detection performance.
Thus, we proposed \name~leveraging the strengths of LiDAR and RGB for proposal generation and then symmetrically using them in the refinement stage.
Our evaluation verifies that our novel approach outperforms the current state-of-the-art significantly for long range 3D detection of pedestrians.

We introduced a new evaluation methodology on KITTI 3D object detection to gain more insights into pedestrian detectors.
Using this methodology, we showed that an architecture with a lower overall KITTI BEV score can still be far superior in long range detection than the best performing approaches.
Due to the importance of long range detection for comfort and safety of autonomous vehicles, we encourage authors to test their performance not only on the overall metric but to separate the metric into different distance ranges.







\section*{ACKNOWLEDGMENT}

The research leading to these results is funded by the German Federal Ministry for Economic Affairs and Energy within the project ”KI-Absicherung” (grant: 19A19005U).

\bibliography{egbib}
\bibliographystyle{ieeetr}

\end{document}